# SurgeryLSTM: A Time-Aware Neural Model for Accurate and Explainable Length of Stay Prediction After Spine Surgery

Ha Na Cho, MS[1], Sairam Sutari, MS[1], Alexander Lopez, MD, MS[2], Hansen Bow, MD, PhD[2], Kai Zheng, PhD[1]

[1]University of California Irvine, Irvine, CA, United States
[2]Department of Neurosurgery, University of California, Irvine, Orange, California, United States

Corresponding author: Ha Na Cho, 6095 Donald Bren Hall, Irvine, CA, 92697-3440, chohn1@uci.edu

**Lay Summary**

Knowing how long a patient will stay in the hospital after elective spine surgery is important for both clinical care and hospital planning. This study developed a novel AI-based method called SurgeryLSTM to predict hospital length of stay (LOS) using data collected before and during surgery. Unlike older models that look at a patient's data all at once, this model looks at changes over time to better understand the patient's health journey. SurgeryLSTM uses a deep learning model called a bidirectional long short-term memory (BiLSTM) network, which allows it to consider both past and future health information in a sequence. It also uses an attention mechanism that highlights the most important time points in the data. This helps the model focus on the moments that matter most for predicting LOS. To make the model's decisions more understandable, the study included explainable AI tools, such as SHAP values, which show which medical factors influenced each prediction. For instance, patients with complex surgeries or chronic conditions like kidney disease were predicted to have longer stays. This research shows how AI can be used to support personalized patient care and hospital resource planning in a transparent and reliable way.

**Abstract**

**Objective**: To develop and evaluate machine learning (ML) models for predicting length of stay (LOS) in elective spine surgery, with a focus on the benefits of temporal modeling and model interpretability.

**Materials and Methods**: We compared traditional ML models (e.g., linear regression, random forest, support vector machine (SVM), and XGBoost) with our developed model, SurgeryLSTM, a masked bidirectional long short-term memory (BiLSTM) with an attention, using structured perioperative electronic health records (EHR) data. Performance was evaluated using the coefficient of determination ($R^2$), and key predictors were identified using explainable AI.

**Results**: SurgeryLSTM achieved the highest predictive accuracy ($R^2=0.86$), outperforming XGBoost ($R^2 = 0.85$) and baseline models. The attention mechanism improved interpretability by dynamically identifying influential temporal segments within preoperative clinical sequences, allowing clinicians to trace which events or features most contributed to each LOS prediction. Key predictors of LOS included bone disorder, chronic kidney disease, and lumbar fusion identified as the most impactful predictors of LOS.

**Discussion**: Temporal modeling with attention mechanisms significantly improves LOS prediction by capturing the sequential nature of patient data. Unlike static models, SurgeryLSTM provides both higher accuracy and greater interpretability, which are critical for clinical adoption. These results highlight the potential of integrating attention-based temporal models into hospital planning workflows.

**Conclusion**: SurgeryLSTM presents an effective and interpretable AI solution for LOS prediction in elective spine surgery. Our findings support the integration of temporal, explainable ML approaches into clinical decision support systems to enhance discharge readiness and individualized patient care.

**Introduction**

Accurately predicting the LOS for patients undergoing elective spine surgery is essential for optimizing hospital resource management, improving patient outcomes, and reducing costs. Postoperative pain levels, metabolic stability, and functional recovery are factors in spine surgery, which impact discharge readiness[1-3]. Unlike emergency procedures, elective surgeries, where patients are scheduled ahead of their surgery, allow for extensive preoperative assessments, offering more predictable outcomes and creating potential for developing robust predictive models to forecast postoperative recovery and discharge timelines. Nonetheless, achieving reliable and highly accurate LOS prediction results remains a challenge due to the interaction of patient demographics and social determinants of health, pre-existing conditions, and surgical factors that influence the patient recovery.

Traditional statistical models have been widely used for LOS prediction, but they struggle to capture nonlinear relationships in high-dimensional datasets[4]. Further, while conventional methods offer interpretability, they are limited in handling complex interactions within large datasets, reducing their predictive power[5]. As a result, there is a growing interest in leveraging artificial intelligence (AI) and machine learning techniques to enhance LOS prediction accuracy. Recent advancements in AI/ML have demonstrated promise in improving predictive performance by capturing complex patterns in patient data[6]. For example, tree-based models (e.g., Extreme Gradient Boosting (XGBoost)[7], Least Absolute Shrinkage and Selection Operator (LASSO))[8] offer robustness in feature selection; temporal models (e.g., Long Short-Term Memory (LSTM)[9], Temporal Convolutional Network (TCN))[10] are adept at modeling sequential dependencies in patients' journey; and transformer-based models (e.g., BioClinical Bidirectional Encoder Representations from Transformers (BioClinicalBERT))[11] have further expanded predictive capabilities by leveraging contextual information from clinical text. However, despite these improvements, challenges remain for LOS prediction including limited generalizability, data quality issues, and lack of interpretability[12]. Additionally, missing values and imbalanced datasets can introduce biases, reducing the reliability of prediction. Therefore, proper data preprocessing is essential to ensure the model's fairness and stability[13].

This study aimed to address these challenges by developing and evaluating AI/ML-based LOS prediction models specifically tailored for elective spine surgeries. We introduce a novel temporal framework that leverages a bidirectional long short-term memory (BiLSTM) network with an attention mechanism to capture sequential dependencies in structured perioperative clinical data. Unlike conventional statistical models, our approach learns temporal dynamics to enhance both prediction accuracy and interpretability. We implemented a robust preprocessing pipeline to address missing values and class imbalances, ensuring model stability and fairness. To further enhance transparency, we applied explainable AI (XAI) techniques and attention weight visualization to identify clinically meaningful predictors and clarify the influence of key variables on LOS prediction outcomes. This study advances the field by applying a transparent, sequence-aware neural model to elective spine surgery data, offering actionable insights for discharge forecasting, hospital resource planning, and personalized care.

## Methods

### Dataset

This study used a retrospective observational design to develop AI/ML models for forecasting LOS in elective spine surgery patients. The data were retrieved from the electronic health records (EHR) system at the University of California Irvine (UCI) Health, including demographic information, diagnosis (ICD-10 codes), surgical details (surgery dates, surgery codes), and laboratory dataset (laboratory test type, laboratory test dates, test values). The study population consisted of patients at the University of California, Irvine (UCI) Medical Center who underwent spine surgery between 2018 to 2023. Patients were eligible if the age is 18 years or older, underwent spine surgery with the following CPT codes: 22840, 22842, 22876, 22844, 22848, 22845, 22590, 22595, 22600, 22614, 22610, 22612, 22630, 22633, 22846, 22847, 22853, 22854, 22859, 22551, 22552, 22554, 22585, 63081, 63082, 22856, 63050, 63001, 63015, 63003, 63016, 63005, 63017, 63045, 63046, 63047, 63020, 63030, 22558, 22585, 2280, and were admitted through planned surgery. Patients admitted through the emergency department, or outside the period, were excluded. Following these criteria, a total of 2,077 patients met the inclusion criteria. The outcome variable, LOS, was measured in days from the surgery date to the discharge date.

The distribution of LOS outcome provides an overview among elective spine surgery patients demonstrates a right-skewed pattern, with most patients discharged within the first few days postoperatively (Figure 1). The highest frequency of discharges occurs between 0 and 5 days, with a gradual decline as LOS increases. Specifically, 21% of patients were discharged on day 1, 17% on day 3, and 12% on day 5.

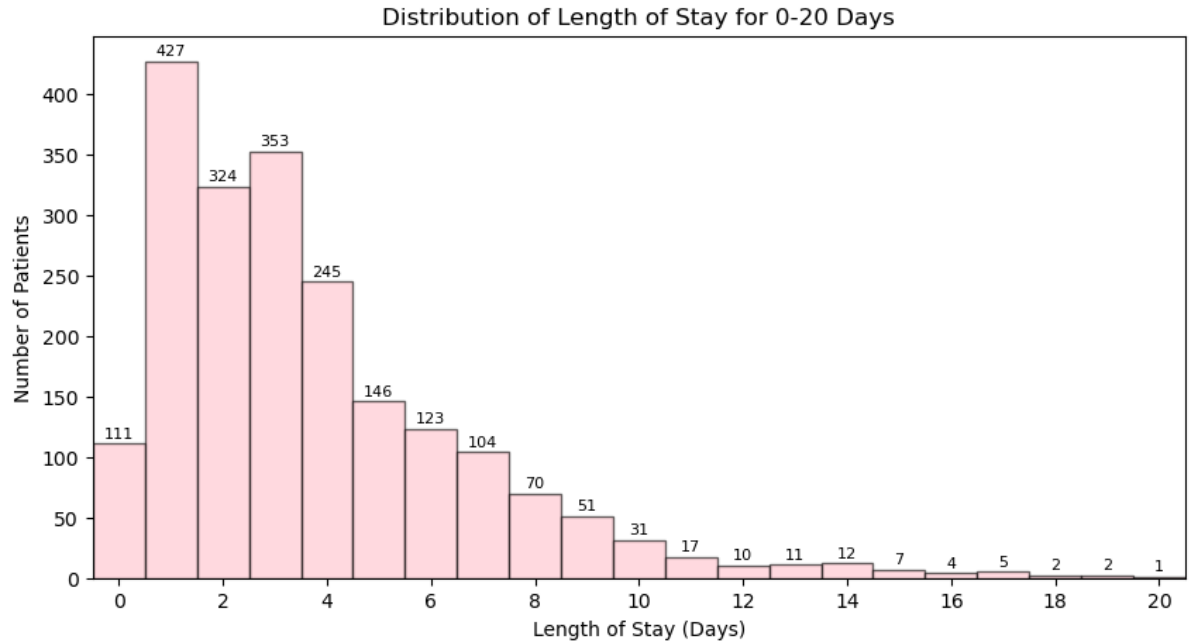

**Figure 1.** Length of stay distribution across patients.

Beyond 10 days, the number of patients significantly decreases, with fewer than 3% of remaining hospitalized beyond two weeks. LOS was limited to 0-20 days, capturing 2,056 of 2,077 patients, while 21 outliers (LOS>20days) were excluded from visualization.

**Data Preprocessing**

To prevent data leakage, all post-surgical information was excluded, retaining only preoperative data for prediction. Temporal variables were normalized relative to the surgery date to ensure consistent chronological alignment. All features were transformed into a structured numeric matrix for compatibility with both ML and LSTM models. Missing values in the dataset arose from inconsistencies in documentation, test availability, and patient variability. A context-specific imputation strategy was applied for lab test results where the majority reference range indicated 'Normal', missing values were imputed using the mean; otherwise, the median was used to reflect distribution patterns more accurately. Lab types with 100% missing values were removed to reduce noise. This ensured that imputed values aligned with clinical relevance while preserving dataset integrity.

A clinical guided approach was used to select the most relevant features for predictive modeling. The 50 most frequently ordered lab tests were selected based on their diagnostic relevance and correlation patterns. With input from two surgeons, 647 features were finalized, incorporating demographics, diagnoses codes, surgery codes, admission and discharge data, and lab results. Diagnosis and surgery codes were standardized using ICD-10 and CPT classifications. Categorical variables were transformed using one-hot encoding, gender was binary encoded (0 = male, 1 = female), and lab results were categorized as 'high', 'low', and 'normal'. Since the dataset primarily consisted of binary and categorical features, standard scaling was selectively applied. Continuous variables, such as age, were scaled using Robust Scaler to minimize the impact of outliers while preserving variability.

Moreover, to handle the class imbalance in LOS categories, a multi-step strategy was implemented. Stratified 5-fold cross-validation was used to ensure balanced representation of LOS categories across training and validation folds. This cross-validation strategy was also employed during hyperparameter tuning, where a randomized search was applied to identify the optimal combination of model parameters based on validation performance. Random oversampling was applied to balance the training dataset by duplicating minority class instances, improving model's learning ability for underrepresented LOS categories. Additionally, class weighting was incorporated during the model's training to penalize misclassification of minority classes, ensuring unbiased predictions.

Temporal variables were processed to maintain the sequential structure required for LSTM modeling. Timestamps (e.g., lab test result date) were converted into relative differences from the surgery date. Only preoperative records were retained to prevent post-surgery information from influencing predictions. The dataset was sorted by patient ID and time to preserve the chronological order of clinical events. For LSTM-based models, time difference was included as a time step identifier.

**Model Architecture and Training**

Our proposed model, surgeryLSTM in Figure 2, leverages a masked Bidirectional LSTM[14] with an attention mechanism to handle variable-length preoperative sequences. Since LSTMs require fixed-length inputs, zero-padding was applied to shorter sequences, and a masking layer ensured the model ignored artificial padding during training. The Bidirectional LSTM captured both past and future dependencies, enhancing the representation of preoperative trends. The attention mechanism dynamically assigned importance weights to different time steps, emphasizing clinically significant events, such as critical lab test results, rather than treating all preoperative records equally. This combination improved both interpretability and predictive performance, allowing the model focused on relevant features influencing LOS.

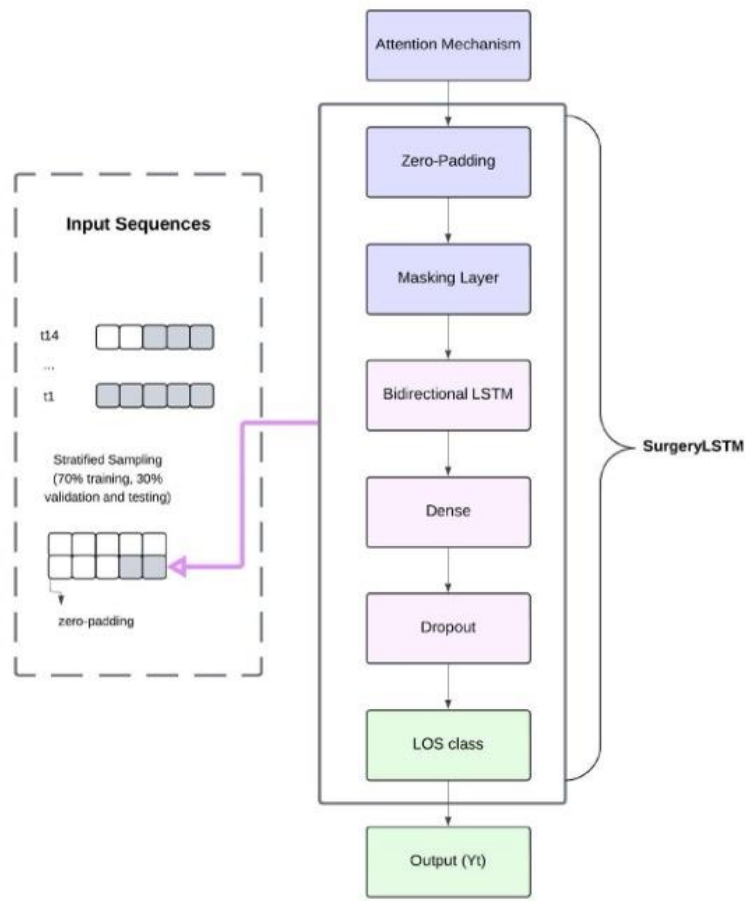

**Figure 2.** SurgeryLSTM model architecture and training workflow.

The dataset was randomly split into 70% training, 30% validation and testing, ensuring stratified sampling to preserve class distribution. Patients had varying numbers of recorded clinical events per patient, thus, a sliding window approach was employed to create fixed-length input sequences while preserving the temporal structure of the data. A sequence length of 14 days (determined based on the 75th percentile of patient records) was chosen to capture sufficient historical trends while preventing excessive padding. For each patient p with $N_p \geq 14$ recorded events, a sequence $S_t$ of length 14 was generated: $S_t = [x_t, x_{t+1}, \ldots, x_{t+13}]$ where $x_t$ represents the feature vector at time step t. The target output was the LOS class, $y_{t+14}$, predicting LOS from the next observed time step. For patients with more than 14 recorded events, overlapping sequences were generated with a stride of 1. Target labels corresponded to the LOS class immediately following the sequence. Hyperparameter tuning was conducted using randomized search with 10 iterations and 3-fold cross-validation. The best set of parameters was selected based on performance metrics, optimizing LSTM units, dense layers, dropout rate, learning rate, batch size, and epochs.

**Model Evaluation**

To benchmark the performance of the LSTM model, multiple machine learning models were trained as baseline comparators, including Linear Regression, Random Forest, Support Vector Machine (SVM), and XGBoost. These models were evaluated using Mean Absolute Error (MAE), Mean Squared Error (MSE), Root Mean Squared Error (RMSE), and Coefficient of Determination ($R^2$) Score. The inclusion of the baselines allowed for a comprehensive comparison of LSTM's performance against traditional ML approaches. Each ML model was selected for its distinct strengths in handling structured clinical data. Linear regression provided an interpretable benchmark while tree-based models such as Random Forest leveraged non-linearity and feature interactions. SVM evaluated distance-based approaches, and XGBoost included for its optimal gradient boosting capabilities. The LSTM model's performance was assessed using the same metrics, ensuring a direct comparison against baseline ML models. Furthermore, to enhance interpretability, SHAP (SHapley Additive exPlanations)[15] force plots were used to analyze individual feature contributions in the XGBoost model.

**Results**

**Patient Demographics**

The study population consisted of 2,077 patients, with a sex distribution of 1,246 males (60.0%) and 831 females (40.0%). The racial composition of the cohort was predominantly White (72.7%), followed by Asian (10.8%), Other Race (9.0%), Black or African American (4.9%), Unknown (1.8%), American Indian or Alaska Native (0.4%), and Native Hawaiian or Other Pacific Islander (0.3%). Most patients fell within the age range of 61-75 years (41.3%), with 21.7% aged 68-75 years. A smaller proportion of patients were younger than 40 years (8.5%), while 9.1% were 82 years or older.

**Model Performance**

**Machine Learning models**

Among the baseline models, XGBoost demonstrated a strong performance achieving an $R^2$ of 0.85, which significantly outperformed linear regression ($R^2 = 0.55$), highlighting the advantages of ensemble methods and instance-based learning for LOS prediction. XGBoost, leveraging its ability to model complex, nonlinear relationships, produced the most accurate predictions. In contrast, linear regression struggled to generalize, informing that LOS prediction requires capturing interactions beyond linear dependencies. Random Forest ($R^2 = 0.60$) and Support Vector Machine ($R^2 = 0.63$) showed moderate performance but lacking the feature selection and gradient optimization strategies that made XGBoost more effective.

      The feature importance analysis from the XGBoost model highlights the key predictors influencing the LOS following elective spine surgery (Figure 3). Bone disorder ranked highest, followed by chronic kidney disease, indicating that underlying skeletal and renal conditions impact hospitalization duration. Other relevant predictors include vitreous disorder, skin abscess, and obesity, suggesting that systemic health conditions contribute to postoperative recovery variations. Procedural factors such as lumbar

decompression, lumbar fusion, anterior fusion, and spinal instrumentation were also strong predictors, reflecting the impact of surgical complexity on LOS. Additionally, cancer metastasis, fatigue, and Body Mass Index (BMI) were moderately influential, showing that patient frailty and metabolic factors affect recovery. To ensure interpretability, ICD-10 and CPT codes were mapped to their clinical counterparts.

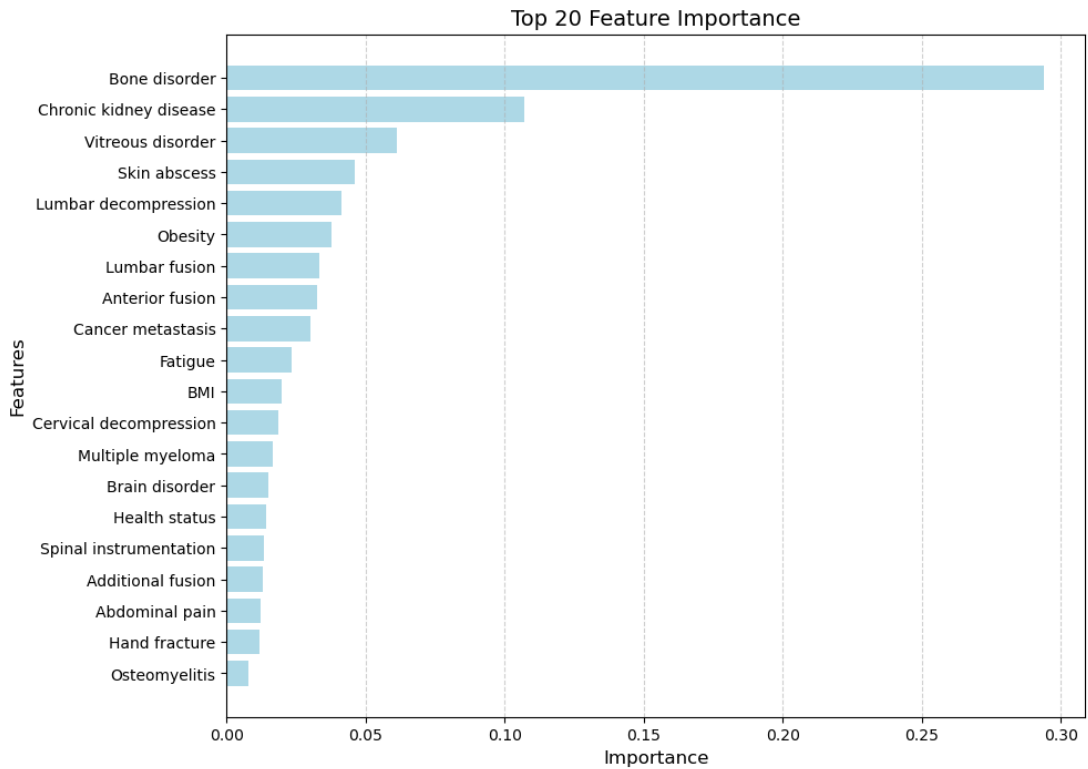

**Figure 3.** Feature importance rankings from the XGBoost model predicting LOS.

**Explainable Artificial Intelligence**

The SHAP summary plot (Figure 4) highlights 20 key predictors influencing LOS, offering a global perspective on model interpretability. These predictors encompass surgical complexity, patient-specific factors, and demographic influences. Surgical procedures, including lumbar decompression, anterior and additional spinal fusion, lumbar spinal fusion, spinal instrumentation, cervical decompression, and lumbar laminectomy, were strongly associated with increased LOS, indicated by higher SHAP (red) shifting rightward. Notably, lumbar decompression showed bidirectional SHAP values, suggesting that for certain patients, this procedure may correspond to shorter stays, potentially due to lower surgical complexity or better baseline health.

Age was another critical determinant, where older patients (red) tend to have shorter LOS, while younger patients (blue) are more likely to experience prolonged hospitalization. Chronic conditions, such as chronic kidney disease, obesity, cancer metastasis, bone disorder, and multiple myeloma, were consistently linked to prolonged LOS, reinforcing the clinical importance of pre-existing comorbidities on post-surgical recovery. Demographic factors, including ethnicity, race, sex, and zip code regions (e.g., 6 and 13), showed mixed directional impacts, indicating the presence of social and structural influences on discharge timing. Additionally, features such as vitreous disorder and general health status contributed modestly, implying broader physiological or overall health influences on recovery.

The SHAP decision plot (Figure 4, right), offers an individualized view of model behavior across 30 randomly selected instances. The visualization confirms that surgical procedures play a dominant role in LOS prediction, with their complexity significantly impacting hospitalization duration. For patients with extended LOS, procedures such as lumber decompression, anterior spinal fusion, and lumbar spinal fusion were primary contributors. Conversely, lower-ranked features, such as brain disorder, and multiple myeloma, appeared to slightly reduce LOS predictions within the observed range, though they do not necessarily correspond to early discharge. This decision plot highlights how combinations of surgical complexity, comorbidities, and demographic variation jointly shape individual outcomes, demonstrating the utility of attention-based modeling in explaining personalized.

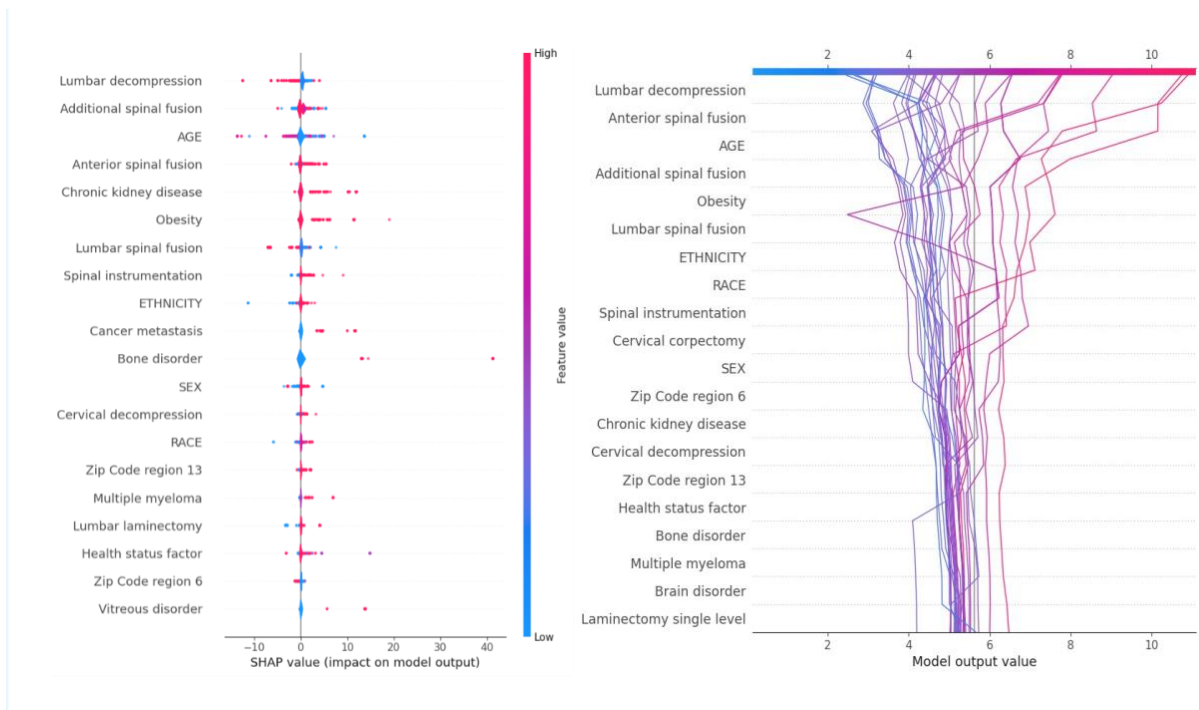

**Figure 4.** SHAP-based model interpretation. The left panel displays a SHAP summary plot showing the top 20 features influencing LOS prediction across the full dataset. Red dots indicate higher feature values; blue dots indicate lower values. The right panel shows a SHAP decision plot for 30 randomly selected patients, visualizing how cumulative feature contributions explain individual predictions.

**LSTM models**

Our model, SurgeryLSTM, incorporating masked BiLSTM with attention mechanism outperformed other traditional ML methods, with lower error rates and higher $R^2$ performance (MAE of 1.48, MSE of 7.21, RMSE of 2.69, and an $R^2$ of 0.86), demonstrating its ability to capture temporal dependencies (Table 1). Three architectural components contributed to its performance: (1) bidirectional processing enhanced model's ability to learn from both past and future temporal dependencies; (2) the attention mechanisms selectively prioritized the most informative segments of patient sequence; and (3) sequence masking improved robustness to variable-length time series. Masked BiLSTM without attention model performed poorly ($R^2 = 0.52$), showing that while bidirectionality helps learning, attention mechanisms are crucial for emphasizing key features and improving the predictive performance.

To further understand model behavior, we examined both learning dynamics and interpretability outputs. The training and validation loss curves show the optimization process of the model over 30 epochs (Figure 5). The loss fluctuates but demonstrates a downward trend, showing effective learning. However, there are continuous spikes in validation loss which suggest the presence of fluctuations in the learning process. The validation loss closely follows the training loss, implying minimal overfitting. Similarly, MAE training trends for training and validation align closely which tells that the model generalizes well to unseen data. The overall decreasing MAE suggests improved error minimization over time.

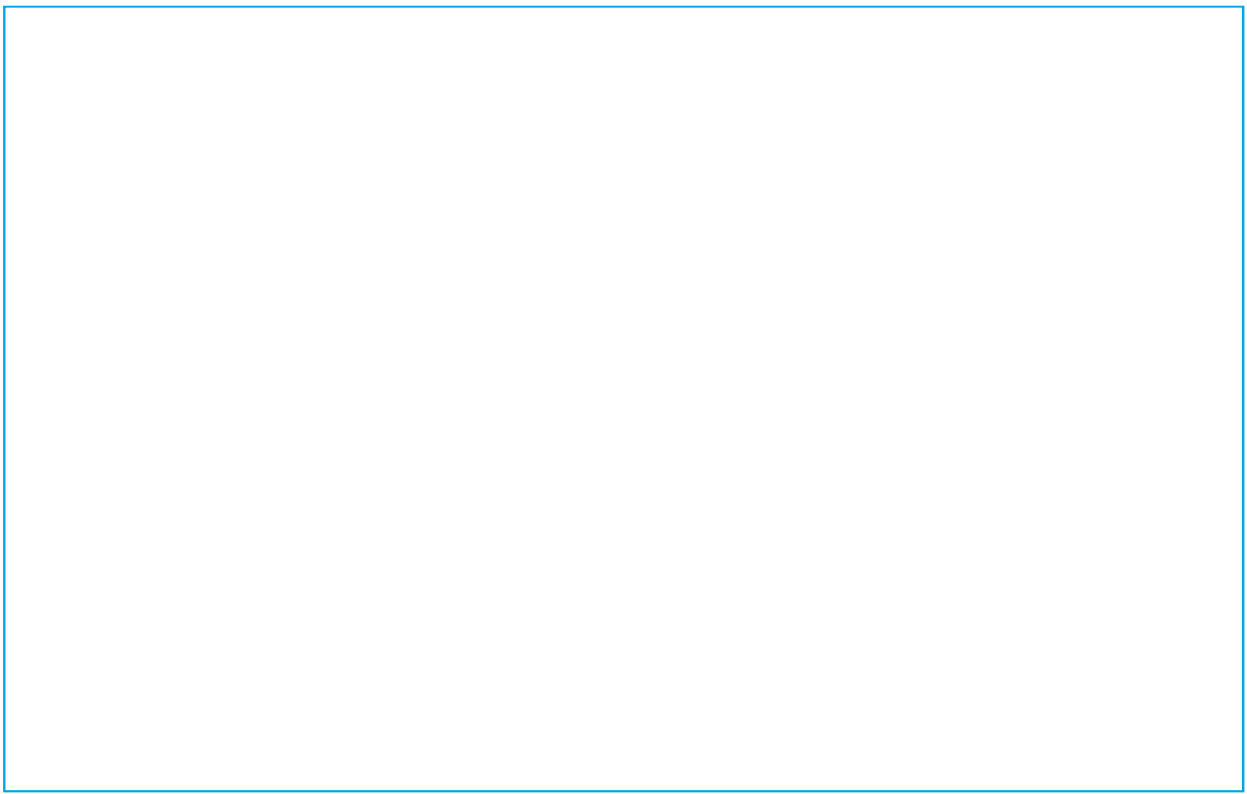

**Figure 5.** Model performance and interpretability diagnostics of the SurgeryLSTM model. Top left/right: Loss and MAE curves across 30 epochs demonstrate effective learning and minimal overfitting. Bottom left: Residuals histogram shows a well-calibrated model with minimal skew. Bottom right: Attention weights peak near the surgery date, highlighting the model's clinical focus.

Crucially, the attention weight distribution provides insights into how the model prioritizes temporal information across the patient timeline. As shown in Figure 5, the model assigns varying levels of importance to time steps, dynamically adjusting focus based on contextual relevance. Higher attention weights are typically concentrated near the time of surgery, indicating that clinical events and observations closer to the procedure are more influential in predicting LOS. This mirrors clinical reasoning, where perioperative status and acute indicators near the surgical event often drive discharge planning. Nonetheless, the model also retains sensitivity to earlier time steps, suggesting that longitudinal trends in patient status (e.g., baseline comorbidities or early lab abnormalities) contribute meaningfully to the prediction.

The accompanying residual distribution histogram (Figure 5) displays the errors centered around zero, indicating good calibration. While most predictions were accurate, a slight left skew in the residual suggests occasional overestimation of LOS in certain patient subgroups, potentially those with atypical recovery trajectories.

| Model Type | Model | MAE | MSE | RMSE | $R^2$ |
|---|---|---|---|---|---|
| **Traditional ML Models** | Linear Regression | 2.68 | 21.99 | 4.69 | 0.55 |
| | Random Forest | 2.69 | 19.64 | 4.43 | 0.60 |
| | Support Vector Machine | 1.07 | 18.45 | 4.30 | 0.63 |
| | XGBoost | 1.63 | 7.59 | 2.75 | 0.85 |
| **LSTM-Based Models** | Masked BiLSTM without attention | 2.73 | 23.73 | 4.87 | 0.52 |
| | **SurgeryLSTM (Masked BiLSTM with attention)** | **1.48** | **7.21** | **2.69** | **0.86** |

Table 1. Comparison of machine learning models for LOS prediction.

**Discussion**

This study demonstrated that SurgeryLSTM achieved the highest predictive performance among all experimented models, highlighting the effectiveness of bidirectional encoding, attention mechanisms, and masking in learning sequential dependencies for LOS prediction. The comparative analysis of traditional ML models and LSTM-based models emphasized the need for effective temporal modeling and preprocessing approach to accurately forecast LOS. While XGBoost performed comparably well, its reliance on static feature relationships limited its ability to capture evolving clinical states. In contrast, SurgeryLSTM leveraged sequential inputs, learning from patient trajectories over time, which improved its capacity to anticipate hospital stays. The ability to integrate both patient health status and surgical complexity contributed to its predictive strength, ensuring that factors such as pre-existing conditions and surgical interventions were dynamically weighted during prediction.

Our findings indicate that LOS is driven by a complex interplay of patient health conditions, surgical complexity, and systemic factors, highlighting the necessity for predictive models that integrate both clinical and procedural elements. The strong influence of chronic conditions such as bone disorder and chronic kidney disease highlights the importance of preoperative risk assessment in optimizing hospital resource allocation and post-surgical management. At the same time, the dominance of surgical features in both SHAP and feature importance analyses suggests that LOS is not solely dictated by pre-existing comorbidities but is also significantly affected by the type and extent of surgical interventions. The presence of mixed SHAP effects for lumbar decompression and age further suggests that individual patient characteristics, surgical decisions, and post-operative care strategies introduce variability in LOS outcomes. These findings reinforce the necessity of developing multi-factorial predictive models that account for both patient-specific risks and procedural complexities, ultimately enhancing hospital workflow efficiency.

A key advantage of SurgeryLSTM was its integration of BiLSTM and an attention mechanism, which enhanced both contextual learning and feature prioritization. Unlike standard LSTMs, which process information unidirectionally, BiLSTM processes sequential data in both forward and backward directions, ensuring that early-stage risk factors are not overshadowed by more recent events. This structure allowed the model to incorporate both historical trends and immediate preoperative conditions, leading to improved generalization. The attention mechanism further refined this process by dynamically assigning greater importance to critical time steps, ensuring that significant preoperative risk factors were weighted more heavily in determining LOS. This is particularly important in hospital settings, where LOS is continuously influenced by a patient's progressing status, requiring a model that can adapt to real-time changes.

Additionally, masking played a crucial role in preventing artificial patterns from emerging due to zero-padding, ensuring that shorter patient sequences did not introduce misleading signals. The performance gap between masked BiLSTM models with and without attention further validated the role of attention in learning clinical dependencies over time. The poor performance of masked BiLSTM without attention indicated that while bidirectional encoding improved temporal representation, it was insufficient for prioritizing key clinical indicators. The attention mechanism bridged this gap by amplifying the influence of time steps most relevant to LOS prediction, leading to enhanced predictive accuracy. Furthermore, attention enabled greater interpretability by illustrating which parts of a patient's clinical history contributed most to the final prediction, allowing clinicians to understand the model's decision-making process, and reinforcing the applicability of LSTM-based approaches for real-time hospital management.

Compared to prior studies in LOS prediction, which often rely on static clinical features or black-box models, our approach offers three distinct innovations several key advancements. Recent studies have explored different methodologies for hospital LOS prediction using traditional ML models to deep learning architectures. Tree-based methods, such as gradient boosting and random forests, have demonstrated strong performance in certain healthcare applications but often struggle with interpretability and fail to capture the evolving nature of clinical states[16]. Transformer-based approaches, such as self-attention mechanisms, have shown promise in predictive modeling for patient cohorts[17]; however, their application to LOS remains limited, with most studies focusing on COVID-19 cohorts rather than elective surgeries. Similarly, convolutional recurrent neural networks have been used to predict hospital stays by leveraging structured and unstructured data[18], but many lack clinical interpretability or are not optimized for sequential structured data. Compared to these approaches, our proposed SurgeryLSTM model extends prior research by integrating BiLSTM with attention mechanisms, allowing dynamic weighting of time steps. Unlike existing deep learning models that treat all time points uniformly, our attention-enhanced LSTMs highlight key clinical moments, thus improving both prediction accuracy and transparency. The use of masked LSTM further mitigates noise from missing data, which is a common limitation in electronic health records[19].

Furthermore, Prior research has explored SHAP analysis for feature interpretation in ML-driven LOS prediction, yet few studies have integrated it with deep learning approaches[20]. By applying SHAP techniques, we identified key predictors such as bone disorder, chronic kidney disease, and lumbar fusion, aligning with previous findings on the impact of comorbidities and surgical complexity on LOS[21]. Our study bridges this gap by combining deep learning and explainable AI for structured HER data in elective spine surgery, a relatively underexplored area. While some studies have applied multimodal ML models to LOS prediction using structured and unstructured data[22], our approach is among the first to incorporate explanations directly into a temporal deep learning framework tailored for elective spine care. This integration advances both predictive accuracy and interpretability, addressing the long-standing challenge of black-box AI models in clinical settings.

Despite its strengths, several practical considerations must be addressed for real-world deployment. While SurgeryLSTM demonstrates strong performance, multi-institutional validation is needed to confirm its generalizability. Integrating structured EHR data with unstructured clinical notes through natural language embeddings may enhance model's ability to capture both static and temporal relationships, improving performance. For real-world impact, models must support real-time inference, ensuring efficiency and seamless integration into clinical workflows. Additionally, hospital systems may face challenges in deploying sequence-based models, including data standardization, interoperability, and clinical training. Embedding LOS predictions into hospital dashboards would enhance usability, while incorporating large language models with structured data could refine risk assessment and improve interpretability.

**Conclusion**

SurgeryLSTM outperforms traditional models by capturing temporal patterns in patient data, offering more accurate and interpretable predictions of length of stay after elective spine surgery. By integrating bidirectional LSTM and attention mechanisms, the model prioritizes clinically relevant time points and adapts to evolving patient conditions. These findings support the use of dynamic, sequence-aware models in real-time clinical decision support.

**Author Contributions**

Ha Na Cho led the study design, developed the methodology, implemented the models, and drafted the manuscript. Sutari Sairam was responsible for data extraction and deidentification from the institutional database. Alexander Lopez provided clinical validation and domain-specific insights. Bow Hansen and Kai Zheng contributed to the study design and provided overall supervision and guidance throughout the research process. All authors contributed to manuscript review and provided critical feedback to improve the final version.

**Funding**

This work did not receive any funding.

**Competing Interests**

There are no conflicts of interest among the authors.

**Data Availability**

The data analyzed in this study were derived from patient health records and contain protected health information. In accordance with HIPAA regulations and institutional review boards policy, the dataset cannot be made publicly available to protect patient privacy. The de-identified data used in the analysis were not approved for public release or external access.

**Ethics Approval**

This study was approved by the Institutional Review Board of the University of California, Irvine (IRB# 4537). All data used were de-identified prior to analysis and complied with relevant ethical guidelines and regulations.